%% file: main.tex
\documentclass[conference]{IEEEtran}
\IEEEoverridecommandlockouts
\usepackage[utf8]{inputenc}
\usepackage{cite}
\usepackage{amsmath,amssymb,amsfonts}
\input{math_commands.tex}

\usepackage{algorithmic}
\usepackage{graphicx}
\graphicspath{{./figs/}}

\usepackage{booktabs}
\usepackage{threeparttable}
\usepackage{colortbl}
\usepackage{xcolor}
\usepackage{color}
\usepackage{tikz}
\usetikzlibrary{shapes,arrows}

\usepackage[hidelinks]{hyperref}
\usepackage{url}
\usepackage{array}
\newcolumntype{H}{>{\setbox0=\hbox\bgroup}c<{\egroup}@{}}

\begin{document}

\title{RPR: Random Partition Relaxation for Training\\Binary and Ternary Weight Neural Networks
\thanks{We would like to thank \emph{Advertima AG} for funding this research. This work has been co-financed by \emph{Innosuisse}.
}
}

\author{
%
\IEEEauthorblockN{Lukas Cavigelli}
\IEEEauthorblockA{ETH Zürich, Switzerland\\
\texttt{cavigelli@iis.ee.ethz.ch}}
\and 
\IEEEauthorblockN{Luca Benini}
\IEEEauthorblockA{ETH Zürich, Switzerland\\
Università di Bologna, Italy\\
\texttt{benini@iis.ee.ethz.ch}}
}

\maketitle

\begin{abstract}
We present Random Partition Relaxation (RPR), a method for strong quantization of neural networks weight to binary (+1/--1) and ternary (+1/0/--1) values. Starting from a pre-trained model, we quantize the weights and then relax random partitions of them to their continuous values for retraining before re-quantizing them and switching to another weight partition for further adaptation.  
We demonstrate binary and ternary-weight networks with accuracies beyond the state-of-the-art for GoogLeNet and competitive performance for ResNet-18 and ResNet-50 using an SGD-based training method that can easily be integrated into existing frameworks.
\end{abstract}


\section{Introduction}
Deep neural networks (DNNs) have become the preferred approach for many computer vision, audio analysis and general signal processing tasks. However, they are also known for their associated high computation workload and large model size. 
As these mostly exceed the capabilities of low-power internet-of-things (IoT) devices, the computation is offloaded to the cloud, but this comes with many drawbacks and potentially prohibitive restrictions: high latency, unreliable connectivity, high operating costs for the compute servers and communication infrastructure, energy-intensive data transmission, and privacy concerns of centralized data processing hinder wide-spread adoption. 

DNN inference close to the sensor where the data is collected is thus en essential step for making IoT and mobile devices smarter and unlock further application scenarios. This requires two aspects to be addressed: computation effort and model size. The first is a major obstacle due to the consequential energy cost and introduced latency, which is often prohibitive for always-on applications. Further, large model sizes are inconvenient for over-the-air updates, drive up the cost of such devices, and on consumer devices, they negatively impact the user experience by taking up lots of storage and prolonging start-up times. 

Many methods have been proposed to address these challenges on multiple levels: 
\begin{itemize}
    \item Optimized network topologies and building blocks such as SqueezeNet \cite{Iandola2016}, MobileNetV2 \cite{Sandler2018}, ShuffleNetV2 \cite{Ma2018}, and Shift \cite{Wu2018a} have fewer trained parameters and require significantly less compute operations for inference \cite{Iandola2017}, and very recent methods also assist in searching for the right network topology and hyperparameters with great success, such as FBNet \cite{Wu2018}. 
    \item Compressing specific networks and/or allow skipping some operations by applying methods such as pruning like EIE \cite{Han2016a} and Deep Compression \cite{Han2016}, or dynamic conditional execution based on the content such as CBinfer \cite{Cavigelli2018,Cavigelli2017} and HD-CNN \cite{Yan2015}. 
    \item Quantizing neural networks can be used to lower the required memory size significantly as well as dramatically simplify the computation and data movement operations. Various methods have been proposed in this area: 1) 8\,bit quantized weights and activations without retraining at negligible accuracy loss \cite{Zhao2019}, 2) binary and ternary weight networks (BWNs, TWNs) with a small accuracy loss after retraining \cite{Courbariaux2015a,Li2016b,Zhu2017}, 3) completely binary and ternary networks with a clear accuracy loss after retraining \cite{Rastegari2016,Darabi2018}, and 4) mixed-precision methods, often in combination with modified topologies \cite{Lin2017}. 
\end{itemize}
Many of these methods can be combined to find an accuracy-energy trade-off point suitable for specific applications, requirements, and target platforms. 

At the same time, recent research into specialized hardware accelerators has shown that improvements by 10--100$\times$ in energy efficiency over optimized software are achievable \cite{Sze2017,Andri2019,Cavigelli2016,Wang2019a}. These accelerators can be integrated into a system-on-chip like those used in smartphones and highly integrated devices for the internet-of-things market. These devices still spend most energy on I/O for streaming data in and out of the hardware unit repeatedly \cite{Cavigelli2019,Cavigelli2018a} as only a limited number of weights can be stored in working memory---and if the weights fit on chip, local memory accesses and the costly multiplications start dominating the energy cost. This allows devices such as \cite{Andri2018a} achieve an energy efficiency of 60\,TOp/s/W for BWN inference even in the mature 65\,nm technology. For comparison, Google's Edge TPU achieves around 2\,TOp/s/W for 8\,bit operations. 

In connection with specialized inference hardware, careful selection of suitable methods is required. Complex network compression schemes are often not compatible with direct hardware inference as decompression is often a lengthy process requiring large memories for dictionaries and a lot of energy. Many recent quantization schemes are now learning the quantization levels to increase the accuracy further, but these commonly do not simplify the arithmetic operations. Further, several strong quantization methods modify and enlarge the network to recover some of the accuracy losses, but with the additional operations, they might not reduce the overall energy cost and model size. 

Strong \emph{and} hardware-friendly quantization of neural networks is crucial to allow more weights to be stored in on-chip working memory, and for them to be loaded more efficiently from external memory. Such quantization also massively simplifies the multiplication operations in the convolution and linear layers, replacing them with lightweight bit-shift operations, or even completely eliminating them in case of binary and ternary weight networks (BWNs, TWNs) \cite{Zhou2017a}. 

In this work, we present a state-of-the-art accuracy method for quantizing the weights of DNNs to binary and ternary values. We formulate the training objective as a mixed-integer non-linear program and train the weights by alternatingly relaxing random partitions of them for optimization while keeping the others quantized. This is compatible with easy integration into standard deep learning toolkits. The paper is organized as follows. We start with an overview over existing strong quantization methods in \secref{sec:relWork}, then introduce our algorithm in \secref{sec:algo}. We proceed to evaluate its performance on several well-known networks in \secref{sec:results}, where we also discuss the results before concluding the paper in \secref{sec:conclusion}.

\section{Related Work}\label{sec:relWork}
Extreme network quantization has started with BinaryConnect \cite{Courbariaux2015a}, proposing deterministic or stochastic rounding during the forward pass and updating the underlying continuous-valued parameters based on the so-obtained gradients, which would naturally be zero almost everywhere. This procedure is also known as the straight-through estimator (STE) in the more general many-bit quantization scenario. 

Then, XNOR-net \cite{Rastegari2016} successfully trained both binary neural networks (BNNs), where the weight and the activations are binarized, as well as BWNs, with a clear jump in accuracy over BinaryConnect by means of dynamic (input-dependent) normalization and for the first time reporting results for a deeper and more modern ResNet topology. 

Shortly after, \cite{Li2016b} presented \emph{ternary weight networks} (TWNs), where they introduced learning the quantization thresholds while keeping the quantization levels fixed and showing a massive improvement over previous work and a top-1 accuracy drop of only 3.6\% on ImageNet, making TWNs a viable approach for practical inference. 

Thereafter, \cite{Zhu2017} introduced \emph{trained ternary quantization} (TTQ), relaxing the constraint of the weights being scaled values of $\{-1,0,1\}$ to $\{\alpha_1,0,\alpha_2\}$. 

A method called \emph{incremental network quantization} (INQ) was developed in \cite{Zhou2017a}, making clear improvements by neither working with inaccurate gradients or stochastic forward passes. Instead, the network parameters were quantized step-by-step, allowing the remaining parameters to adapt to the already quantized weights. This further improved the accuracy of TWNs and fully matched the accuracy of the baseline networks with 5\,bit and above. 

Last year, \cite{Leng2018} presented a different approach to training quantized neural networks by relying on the \emph{alternating direction method of multipliers} (ADMM) more commonly used in chemical process engineering. They reformulated the optimization problem for quantized neural networks with the objective function being a sum of two separable objectives and a linear constraint. ADMM alternatingly optimizes each of these objectives and their dual to enforce the linear constraint. In the context of quantized DNNs, the separable objectives are the optimization of the loss function and the enforcement of the quantization constraint, which results in projecting the continuous values to their closest quantization levels. While ADMM achieves state-of-the-art results to this day, it requires optimization using the extragradient method, thereby becoming incompatible with standard DNN toolkits and hindering wide-spread adoption. 

A few months ago, \emph{quantization networks} (QNs) was introduced in \cite{Yang2019}. They pursue a very different approach, annealing a smoothed multi-step function the hard steps quantization function while using L2-norm gradient clipping to handle numerical instabilities during training. They follow the approach of TTQ and learn the values of the quantization levels. 

Several recent methods, such as HAQ \cite{Wanga} and HAWQ \cite{Dong2019a}, explore mixed-precision quantization of networks for the weights and activations with a different number of quantization levels for each layer. This is aids in further compressing the model size, and such weights can be decompressed on-the-fly in hardware with very little overhead. Our proposed algorithm for quantizing the weights is not restricted to a single set of quantization levels for the entire network and could also be integrated into a mixed-precision training setup. For better comparability to other quantization methods without the influence of the fine-grained selection of the number of quantization levels, we do not focus our evaluations on mixed-precision networks.

\section{RPR: Random Partition Relaxation Training} \label{sec:algo}
In this section, we describe the intuition behind RPR, its key components, and their implementation. 

When training DNNs, we optimize the network's parameters $\vw\in\sR^d$ to minimize a non-convex function $f$,
\begin{align}
    \min_{\vw\in\sR^d} f(\vw). \label{eq:nlp}
\end{align}
This has been widely and successfully approached with stochastic gradient descent-based methods for DNNs in the hope of finding a good local optimum close to the global one of this non-convex function. 

As we further constrain this optimization problem by restricting a subset of the parameters to take value in a finite set of quantization levels $\sL$, we end up with a mixed-integer non-linear program (MINLP): 
\begin{align}
    \min_{\vw_q, \vw_c} f(\vw_q, \vw_c) \qquad \mathrm{s.t.} \quad \vw_q\in\sL^{d_q}, \quad \vw_c\in\sR^{d_c}, \label{eq:minlp}
\end{align}
where $\vw_q$ are the quantized (e.g., filter weights) and $\vw_c$ the continuous parameters (e.g., biases, batch norm factors) of the network. Common sets of quantization levels $\sL$ are symmetric uniform with or without zero ($\{0\}\cup\{\pm i\}_i$ or $\{\pm i\}_i$) and symmetric exponential ($\{0\}\cup\{\pm 2^i\}_i$) due to their hardware suitability (multiplications can be implemented as bit-shifts). Less common but also used are trained symmetric or arbitrary quantization levels ($\{\pm \alpha_i\}_i$ or $\{\alpha_i\}_i$). Typically, the weights of the convolutional and linear layers are quantized except for the first and last layers in the network, since quantizing these has been shown to have a much stronger impact on the final accuracy than that of the other layers \cite{Li2017d,Rastegari2016}. As in most networks the convolutional and linear layers are followed by batch normalization layers, linear scaling of the quantization levels has no impact on the optimization problem. 

Mixed-integer non-linear programs such as \eqref{eq:minlp} are NP-hard and practical optimization algorithms trying to solve it are only approximate. Most previous works approach this problem by means of annealing a smoothed multi-step function applied to underlying non-quantized weights (and clipping the gradients) or by quantizing the weights in the SGD's forward pass and introducing proxy gradients in the backward pass (e.g., the straight-through estimator (STE)) to allow the optimization to progress despite the gradients being zero almost everywhere. Recently, \cite{Leng2018} proposed to use the \emph{alternating direction method of multipliers (ADMM)} to address this optimization problem with promising results. However, their method requires a non-standard gradient descent optimizer, thus preventing simple integration into commonly used deep learning toolkits and thereby wide-spread adoption.

\subsection{Random Partition Relaxation Algorithm}
For RPR, we propose to approach the MINLP through alternating optimization. 
Starting from continuous values for the parameters in $\sW_q$, we randomly partition $\sW_q$ into $\sW_q^\mathrm{constr}$ and $\sW_q^\mathrm{relaxed}$ for some specified freezing fraction (FF), e.g. $\mathrm{FF}=\frac{\#\sW_q^\mathrm{constr}}{\#\sW_q}=90\%$. The parameters in $\sW_q^\mathrm{constr}$ are quantized to their closest value in $\sL$ while those in $\sW_q^\mathrm{relaxed}$ keep their continuous value, which is updated according to
\begin{align}
    \hat{\vw}_q^\mathrm{relaxed}, \hat{\vw}_c = \argmin_{\vw_q^\mathrm{relaxed}, \vw_c} f(\vw_q^\mathrm{constr}, \vw_q^\mathrm{relaxed}, \vw_c). \label{eq:altOptim}
\end{align}
\begin{figure*}[tb]
    \centering
    \includegraphics[width=0.92\linewidth]{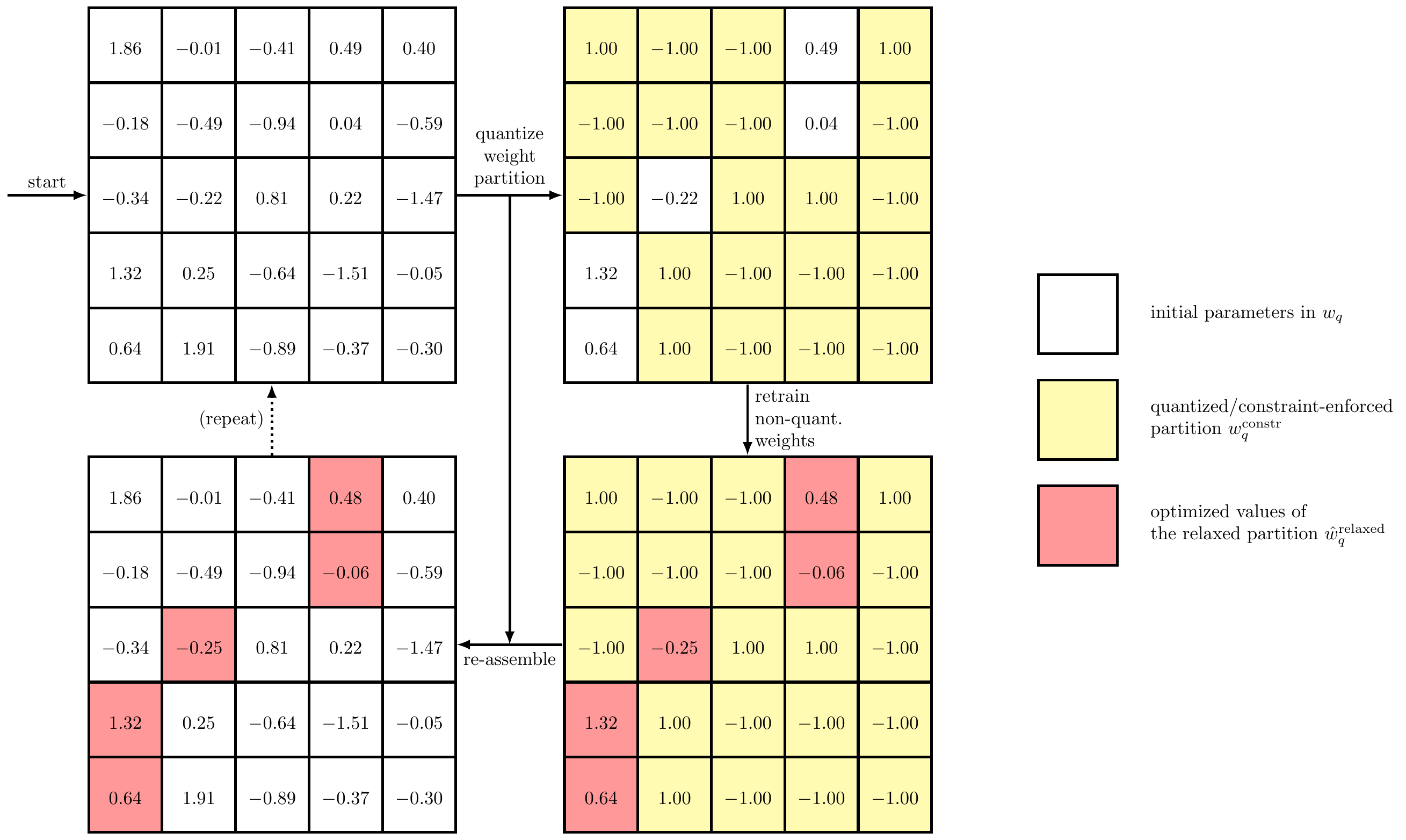}
    \caption{Overview of the Random Partition Relaxation (RPR) algorithm.}
    \label{fig:algoFlow}
\end{figure*}

This allows the relaxed parameters to co-adapt to the constrained/quantized ones. This step is repeated, alternating between optimizing other randomly relaxed partitions of the quantized parameters (cf. \figref{fig:algoFlow}). As the accuracy converges, $\mathrm{FF}$ is increased until it reaches 1, at which point all the constrained parameters are quantized. 

The non-linear program \eqref{eq:altOptim} can be optimized using standard SGD or its derivatives like Adam, RMSprop. We have experimentally found performing gradient descent on \eqref{eq:altOptim} for one full epoch before advancing to the next random partition of $\sW_q$ to converge faster than other configurations. Note that $\vw_q^\mathrm{constr}$ is always constructed from the underlying continuous representation of $\vw_q$. We also initialize $\vw_q^\mathrm{relaxed}$ to the corresponding continuous-valued representation as well, thus providing a warm-start for optimizing \eqref{eq:altOptim} using gradient descent.

\subsection{Initialization}
Starting with the standard initialization method for the corresponding network has worked well for training VGG-style networks on CIFAR-10 and ResNet-18 on ImageNet. We experimentally observed that smaller freezing fractions $\mathrm{FF}$ can be used for faster convergence at the expense of less reliable convergence to a good local optimum. 

However, a network can be quantized much faster and tends to reach a better local optimum when starting from a pre-trained network. When convolution and linear layers are followed by a batch normalization layer, their weights become scale-invariant as the variance and mean are immediately normalized, hence we can define our quantization levels over the range $[-1,1]$ without adding any restrictions. However, the continuous-valued parameters of a pre-trained model might not be scaled suitably. We thus re-scale each filter of each layer $i$ to minimize the $\ell_2$ distance between the continuous-valued and the quantized parameters, i.e. 
\begin{gather}
    \widetilde{\vw}^{(i)} = \frac{1}{\hat{s}^{(i)}}\vw^{(i)}\label{eq:init1}\\
         \mathrm{with}\quad
    \hat{s}^{(i)} = \argmin_{s\ge0} \| \vw^{(i)}-s\vw^{(i)}_\mathrm{quant} \|_2 \label{eq:init2}\\
        \mathrm{and}\quad
    \vw^{(i)}_\mathrm{quant} = \argmin_{\ell\in\sL} |\vw^{(i)}-\ell|.
        \label{eq:init3}
\end{gather}

Practically, we implemented \eqref{eq:init2} using a grid search for $s$ over 1000 points spread uniformly over $[0,\max_i|w_i|]$ before locally fine-tuning the best result using the downhill simplex method. The time for this optimization is negligible relative to the overall compute time and in the range of a few minutes for all the weights to be quantized within ResNet-50. 


\section{Experimental Results}
We conducted experiments on ImageNet with ResNet-18, ResNet-50, and GoogLeNet in order to show the performance of RPR by training them as binary weight and ternary weight networks. We refrain from reporting results on CIFAR-10 and with AlexNet on Imagenet as these networks are known to be over-parametrized and thus rely on additional regularization techniques not to overfit---this is an irrelevant scenario for resource-efficient deployment of DNNs, as a smaller DNN would be selected anyway. Following common practice, we do not quantize the first and last layers of the network. If not stated otherwise, we start from the corresponding pre-trained model available through the torchvision v0.4.0 library. We ran the training for each configuration only once and did not perform cherry-picking across multiple runs. 

\subsection{Preprocessing}
The preprocessing and data augmentation methods used in related work vary wildly and from simple image rescaling and cropping with horizontal flips and mean/variance normalization to methods with randomized rescaling, cropping to different aspect ratios, and brightness/contrast/saturation/lighting variations. Consistent with literature, we have found that a quite minimal preprocessing by rescaling the image such that the shorter edge has 256\,pixels followed by random crops of $224\times224$ pixels and random horizontal flips showed best results. During testing, the same resizing and a $224\times224$ center crop were applied. 
We observed simpler preprocessing methods working better: this is expected as the original networks' capacities are reduced by the strong quantization, and training the network to correctly classify images sampled from a richer distribution of distortions than that of the original data takes away some of the capacity of the network.

\subsection{Hyperparameter Selection \& Retraining Time}
We trained the networks using the Adam optimizer with initial learning rates identical to the full-precision baseline models ($10^{-3}$ for all models). Over the time of the training procedure, the fraction of frozen/quantized weights has to increase to 1 in order to complete the quantization procedure. Starting from a pre-trained network yielded better final results than randomized initialization. During an initial quantization-aware training phase, we use a freezing fraction of $\mathrm{FF}=0.9$ until stabilization of the validation metric. This initial $\mathrm{FF}$ is quite robust---we have observed identical results for $\mathrm{FF}\in[0.75,0.925]$. Leaving this range leads to worse final results, where choosing the highest $\mathrm{FF}$ without adverse effects is generally desirable for training speed (fewer steps needed for $\mathrm{FF}$ to converge to 1) and higher values seemed to hinder optimization. 

\begin{figure}[tb]
    \centering
    \includegraphics[width=\linewidth]{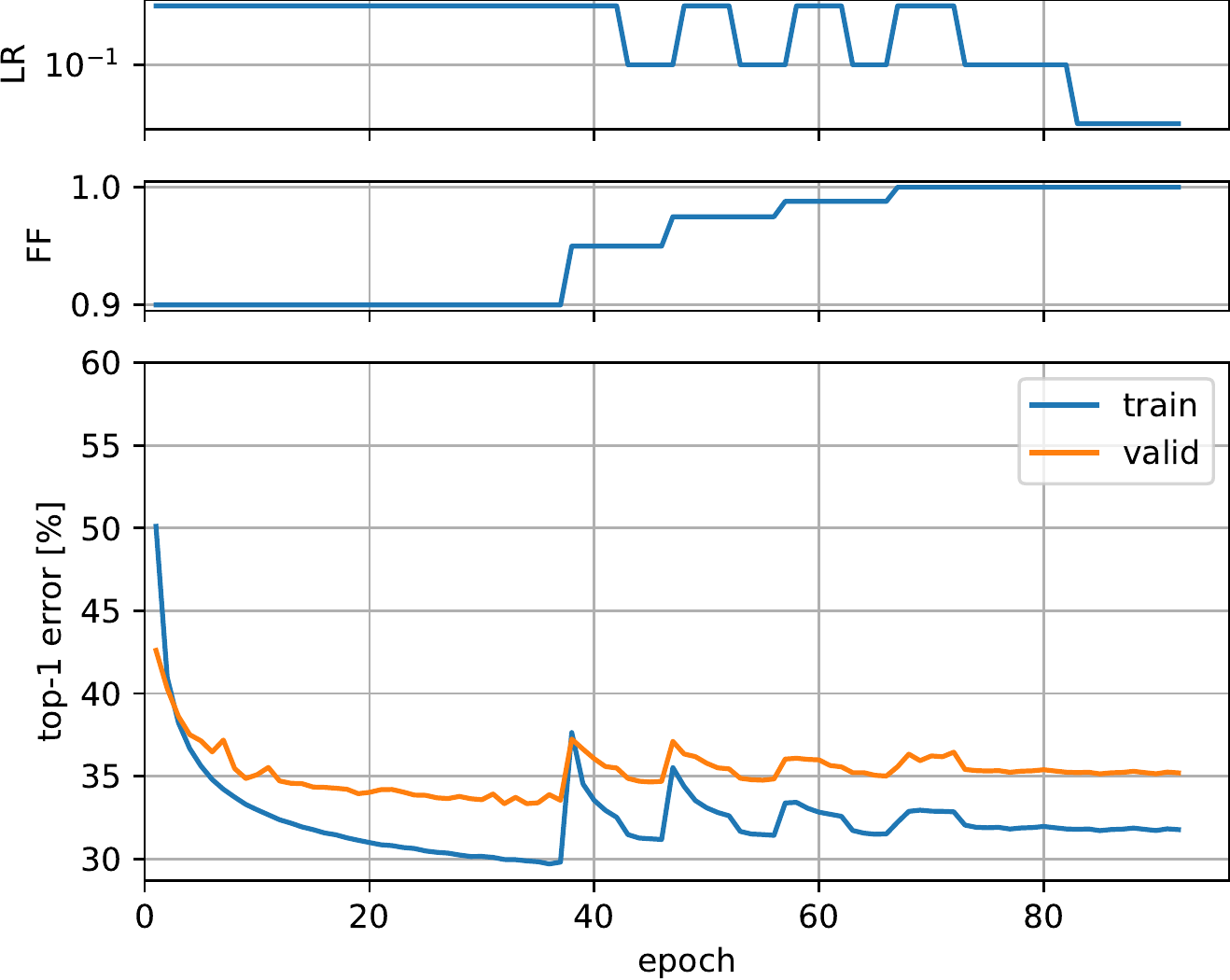}
    \caption{Evolution of the top-1 training and test accuracy together with the schedules for the freezing fraction (FF) and the learning rate (LR) while re-training GoogLeNet as a ternary weight network.}
    \label{fig:learningCurve}
\end{figure}

We proceed with cutting the size of the partition of relaxed weights in half three times before freezing all the weights. Specifically, we went to $\mathrm{FF}=0.95, 0.975, 0.9875, 1.0$. Each different FF was kept for 15 epochs, always starting with the initial learning rate and reducing it by $10\times$ after 10 epochs at the specific FF. Waiting a larger number of epochs before reducing the learning rate or increasing $\mathrm{FF}$ has shown no significant effect on the final accuracy. After reaching $\mathrm{FF}=1.0$, the learning rate is kept for 10 cycles each at $1\times$, $0.1\times$, and $0.01\times$ the initial learning rate, which still fine-tunes the batch normalization parameters. An example of a freezing fraction and learning rate schedule is shown in \figref{fig:learningCurve}. 

In practice, quantizing a network with RPR requires a number of training epochs similar to training the full-precision model, and there is no significant difference in execution time relative to non-quantized training epoch. This is shown for the quantization of GoogLeNet to ternary weights in \figref{fig:learningCurve}. The quantization with $\mathrm{FF}=0.9$ requires 37 epochs followed by 45 epochs of iteratively increasing FF before a final phase of optimizing only the continuous parameters for 30 additional epochs.

\subsection{Results \& Comparison} \label{sec:results}

\begin{table*}[tb]
    \centering
    \caption{Experimental Results on ImageNet}
    \label{tab:imageNetResults}
    \begin{threeparttable}
    \begin{tabular}{lllHlcc}
        \toprule
        Model & Method\tnote{$\star$} &  & Preproc. & Levels\tnote{$\dagger$} & \multicolumn{2}{c}{------ Accuracy ------} \\ 
         &  &  &  &  & top-1 [\%] & top-5 [\%] \\ 
        \midrule
        ResNet-18  &         baseline & \ torchvision v0.4.0 &  & full-prec. & 69.76 & 89.08 \\ 
        ResNet-18  &  QN & \cite{Yang2019} & A & 5: $\{\alpha_i\}_i$ & 69.90 & 89.30 \\
        ResNet-18  &             ADMM & \cite{Leng2018} &  & 5: $\{0\}\cup\{\pm2^i\}_i$  & 67.50 & 87.90 \\
        ResNet-18  &          LQ-Nets & \cite{Zhang2018a} &  & 4: $\{\pm\alpha_i\}_i$  & 68.00 & 88.00 \\ 
        ResNet-18  &  QN & \cite{Yang2019} & A & 3: $\{\alpha_1,\alpha_2,\alpha_3\}$ & 69.10 & 88.90 \\
        ResNet-18+\tnote{$\ddagger$} &  TTQ & \cite{Zhu2017} &  & 3: $\{\alpha_1,0,\alpha_2\}$  & 66.60 & 87.20 \\
        ResNet-18  &             ADMM & \cite{Leng2018} & C & 3: $\{-1,0,1\}$ & 67.00 & 88.00 \\
        ResNet-18  &              INQ\tnote{§} & \cite{Zhou2017a} & unknown & 3: $\{-1,0,1\}$ & 66.00 & 88.00 \\
        ResNet-18+\tnote{$\ddagger$} &   TWN & \cite{Li2016b} &  & 3: $\{-1,0,1\}$ & 65.30 & 86.20 \\ 
        ResNet-18  &   TWN & \cite{Li2016b} &  & 3: $\{-1,0,1\}$ & 61.80 & 84.20 \\
        ResNet-18  & \textbf{RPR (ours)} &  &  & 3: $\{-1, 0, 1\}$  & \textbf{66.31} & \textbf{87.84} \\
        ResNet-18  &             ADMM & \cite{Leng2018} & C & 2: $\{-1, 1\}$ & 64.80 & 86.20 \\
        ResNet-18  &     XNOR-net BWN & \cite{Rastegari2016} &  & 2: $\{-1, 1\}$ & 60.80 & 83.00 \\
        ResNet-18  & \textbf{RPR (ours)} &  &  & 2: $\{-1, 1\}$  & \textbf{64.62} & \textbf{86.01} \\
        \midrule
        ResNet-50  &         baseline & \ torchvision v0.4.0 &  & full-prec. & 76.15 & 92.87 \\ 
        ResNet-50  &             ADMM & \cite{Leng2018} & C & 3: $\{-1,0,1\}$ & 72.50 & 90.70 \\
        ResNet-50  &   TWN & \cite{Li2016b} &  & 3: $\{-1,0,1\}$ & 65.60 & 86.50 \\
        ResNet-50  & \textbf{RPR (ours)} &  &  & 3: $\{-1, 0, 1\}$  & \textbf{71.83} & \textbf{90.28} \\
        ResNet-50  &             ADMM & \cite{Leng2018} & C & 2: $\{-1, 1\}$ & 68.70 & 88.60 \\
        ResNet-50  &     XNOR-net BWN & \cite{Rastegari2016} &  & 2: $\{-1, 1\}$ & 63.90 & 85.10 \\
        ResNet-50  & \textbf{RPR (ours)} &  &  & 2: $\{-1, 1\}$  & \textbf{65.14} & \textbf{86.31} \\
        \midrule
        GoogLeNet  &         baseline & \ torchvision v0.4.0 &  & full-prec. & 69.78 & 89.53 \\
        GoogLeNet  &             ADMM & \cite{Leng2018} & C & 3: $\{-1,0,1\}$ & 63.10 & 85.40 \\
        GoogLeNet  &   TWN & \cite{Li2016b} &  & 3: $\{-1,0,1\}$ & 61.20 & 84.10 \\
        GoogLeNet  & \textbf{RPR (ours)} &  &  & 3: $\{-1, 0, 1\}$  & \textbf{64.88} & \textbf{86.05} \\
        GoogLeNet  &             ADMM & \cite{Leng2018} & C & 2: $\{-1, 1\}$ & 60.30 & 83.20 \\
        GoogLeNet  &     XNOR-net BWN & \cite{Rastegari2016} &  & 2: $\{-1, 1\}$ & 59.00 & 82.40 \\
        GoogLeNet  & \textbf{RPR (ours)} &  &  & 2: $\{-1, 1\}$  & \textbf{62.01} & \textbf{84.83} \\ 
        \bottomrule
    \end{tabular}
    \begin{tablenotes}
        \item [$\star$] Unless noted otherwise, the ResNet models have Type-B bypasses (with a $1\times 1$ convolution in the non-residual paths on increase of the feature map count). 
        \item [$\dagger$] Unless noted otherwise, the first and last layers are excluded from quantization. 
        \item [$\ddagger$] Modified network: each layer has $2.25\times$ as many weights.
        \item [§] First and last layers are also quantized. 
    \end{tablenotes}
    \end{threeparttable}
\end{table*}
We provide an overview of our results and a comparison to related work in Table~\ref{tab:imageNetResults}. For ResNet-18, our method shows similar accuracy to the ADMM-based method, clearly outperforming other methods such as the XNOR-net BWN, TWN, and INQ. As discussed before, the ADMM algorithm requires an optimization procedure that is not a simple variation of SGD and has thus not yet found wide-spread adoption. Specifically, the authors did not release code to reproduce the results or the trained models, and to the best of our knowledge no public re-implementation is available. Unfortunately, no indication of the required training time has been reported for ADMM.

A higher accuracy than RPR is achieved by TTQ with an enlarged network ($2.25\times$ as many parameters) and by \emph{Quantization Networks}. Both methods however, introduce trained quantization levels with dire consequences for hardware implementations: either as many multipliers as in full-precision networks are required, or the operations are transformed as $\sum_i w_i x_i = \alpha_1 \sum_{i}\sOne_{w_i=\alpha_1}x_i + \alpha_2 \sum_{i}\sOne_{w_i=\alpha_2}x_i + \alpha_3 \sum_{i}\sOne_{w_i=\alpha_3}x_i$, requiring only very few multiplications but 3 adder trees, thereby increasing the required silicon area for the main compute logic by $\approx3\times$ with respect to a TWN with a fixed set of quantization levels $\sL=\{-1,0,1\}$ and can thus be expected to have corresponding effects on energy. For ResNet-50, the results look similar: we achieved accuracies close to the state-of-the-art (i.e., ADMM), but avoiding the added complexity of altering the optimization method beyond a simple derivative of SGD. 

For GoogLeNet we surpass the current state-of-the-art, ADMM, by 1.7\% top-1 accuracy for binary weights and 1.76\% for ternary weights. 

\section{Conclusion} \label{sec:conclusion}
We have proposed using alternating optimization for training strongly weight-quantized neural networks by randomly relaxing the quantization constraint on small fractions of the weights. We have implemented this method using standard SGD-based optimization. This method improves the state-of-the-art accuracy for binary and ternary weight GoogLeNet and achieves accuracies similar to previous methods on ResNet-18 and ResNet-50 while maintaining easy integrability into existing deep learning toolkits by using standard gradient descent optimization. 

\bibliographystyle{IEEEtran}
\bibliography{references}

\end{document}

%% file: math_commands.tex
\usepackage{amsmath,amsfonts,bm}
\usepackage{dsfont}

\def\figref#1{Figure~\ref{#1}}

\def\secref#1{Section~\ref{#1}}

\def\eqref#1{(\ref{#1})}

\def\1{\bm{1}}

\def\vw{{\bm{w}}}

\DeclareMathAlphabet{\mathsfit}{\encodingdefault}{\sfdefault}{m}{sl}
\SetMathAlphabet{\mathsfit}{bold}{\encodingdefault}{\sfdefault}{bx}{n}

\def\sOne{{\mathds{1}}}

\def\sL{{\mathbb{L}}}

\def\sR{{\mathbb{R}}}

\def\sW{{\mathbb{W}}}

\DeclareMathOperator*{\argmin}{arg\,min}